\pdfoutput=1

\documentclass[sigconf]{acmart}

\AtBeginDocument{%
  \providecommand\BibTeX{{%
    \normalfont B\kern-0.5em{\scshape i\kern-0.25em b}\kern-0.8em\TeX}}}

\usepackage{xcolor,colortbl}

\usepackage{tikz}
\usepackage{pgfplots}
\usetikzlibrary{patterns}

\copyrightyear{2022}
\acmYear{2022}
\setcopyright{rightsretained}
\acmConference[FAccT '22]{2022 ACM Conference on Fairness, Accountability, and Transparency}{June 21--24, 2022}{Seoul, Republic of Korea}
\acmBooktitle{2022 ACM Conference on Fairness, Accountability, and Transparency (FAccT '22), June 21--24, 2022, Seoul, Republic of Korea}\acmDOI{10.1145/3531146.3533183}
\acmISBN{978-1-4503-9352-2/22/06}

\begin{document}

\title{Markedness in Visual Semantic AI}

\author{Robert Wolfe}
\affiliation{%
  \institution{University of Washington}
  \city{Seattle}
  \state{WA}
  \country{USA}}
\email{rwolfe3@uw.edu}

\author{Aylin Caliskan}
\affiliation{%
  \institution{University of Washington}
  \city{Seattle}
  \state{WA}
  \country{USA}}
\email{aylin@uw.edu}

\begin{abstract}
We evaluate the state-of-the-art multimodal "visual semantic" model CLIP ("Contrastive Language Image Pretraining") for biases related to the  marking of age, gender, and race or ethnicity. Given the option to label an image as "a photo of a person" or to select a label denoting race or ethnicity, CLIP chooses the "person" label 47.9\% of the time for White individuals, compared with 5.0\% or less for individuals who are Black, East Asian, Southeast Asian, Indian, or Latino or Hispanic. The model is also more likely to rank the unmarked "person" label higher than labels denoting gender for Male individuals (26.7\% of the time) vs. Female individuals (15.2\% of the time). Age also affects whether an individual is marked by the model: Female individuals under the age of 20 are more likely than Male individuals to be marked with a gender label, but less likely to be marked with an age label, while Female individuals over the age of 40 are more likely to be marked based on age than Male individuals. We trace our results back to the CLIP embedding space by examining the self-similarity (mean pairwise cosine similarity) for each social group, where higher self-similarity denotes greater attention directed by CLIP to the shared characteristics (\textit{i.e.,} age, race, or gender) of the social group. The results indicate that, as age increases, the self-similarity of representations of Female individuals increases at a higher rate than for Male individuals, with the disparity most pronounced at the "more than 70" age range. Six of the ten least self-similar social groups are individuals who are White and Male, while all ten of the most self-similar social groups are individuals under the age of 10 or over the age of 70, and six of the ten are Female individuals. Our results yield evidence that bias in CLIP is intersectional: existing biases of self-similarity and markedness between Male and Female gender groups are further exacerbated when the groups compared are individuals who are White and Male and individuals who are Black and Female. CLIP is an English-language model trained on internet content gathered based on a query list generated from an American website (Wikipedia), and results indicate that CLIP reflects the biases of the language and society which produced this training data.
\end{abstract}



\maketitle

\section{Introduction}

Recent progress in multimodal artificial intelligence (AI) has produced the first "zero-shot" language-and-vision model, known as CLIP ("Contrastive Language Image Pretraining"), which allows for the definition of image classes in natural language and achieves performance competitive with the state of the art on datasets on which it was not trained \cite{radford2021learning}. The advances achieved by CLIP have spawned a new wave of interest and innovation in "visual semantic" AI, which combines language and image representations in the same embedding space. The past year has seen the development of numerous proprietary models similar to CLIP \cite{jia2021scaling,tiwary_2021}, which are likely to serve as important components in the future foundation of the internet \cite{nayak_2021}. Like word embeddings and language models, visual semantic AI is trained using human-authored language on an internet-scale dataset: CLIP's training data is composed of 400 million pairs of images and associated text scraped from the English-language internet \cite{radford2021learning}. Prior work has uncovered semantic biases in CLIP associating immigrants and religious minorities with unpleasant stereotypes \cite{goh2021multimodal}, and gender biases related to underrepresentation of women when using the model for image retrieval \cite{wang2021gender}. In this research, we systematically evaluate CLIP for a previously unobserved perceptual bias: the bias of who is marked, and based on what socially defined characteristics (gender, race, and age).

Our contributions are twofold. First, we find disparities in the self-similarity (mean pairwise cosine similarity) of embedded images along the axes of gender, age, and race or ethnicity. Concretely, embedded images of White individuals in the FairFace dataset have self-similarity of 0.573, compared to self-similarity of $>.62$ for Indian, East Asian, or Southeast Asian individuals; images of Male individuals have mean self-similarity of 0.566, compared to 0.592 for images of Female individuals; and images of individuals between 40 and 49 have mean cosine similarity of 0.573, compared to 0.623 for individuals more than 70 and 0.692 for individuals between 0 and 2. These results are also intersectional: of 126 social groups formed based on the intersection of gender, age, and race or ethnicity, the ten least self-similar social groups are all Male, and six of the ten are White. The ten most self-similar groups all reflect people either under the age of 10 or over the age of 70, and six of the ten are Female.

Second, we show that CLIP prefers to label images of members of some social groups based on demographic characteristics such as race, gender, and age, and to omit those labels when describing other social groups. We prompt CLIP to rank the text label "a photo of a person" against sets of text labels related to gender, age, and race or ethnicity based on the FairFace dataset \cite{karkkainen2021fairface}. If the model ranks "a photo of a person" higher than any of the FairFace gender labels, we consider that image to be unmarked based on gender. We repeat the same experiment using the set of age labels and the set of race or ethnicity labels from FairFace. CLIP ranks the "a photo of a person" label highest 47.9\% of the time for images of White individuals, and less than 16.9\% of the time for all other racial or ethnic groups; CLIP ranks "a photo of a person" highest 26.6\% of the time for images of Male individuals, and 15.2\% of the time for Female individuals; and CLIP ranks "a photo of a person" highest 59.0\% of the time for images of individuals between 30 and 39, but only 6.7\% of the time for images of individuals older than 70. Differences are also intersectional, as Female individuals are marked according to their gender at younger ages than Male individuals, and Female individuals who are more than 40 years old are more likely to be marked based on age than Male individuals at the same age. The disparity is greater in almost all cases between individuals who are White and Male and individuals who are Black and Female.

This form of AI bias is likely to become increasingly important as multimodal AI models use natural language supervision \cite{radford2021learning} to unify different forms of information, such as vision, language, speech, and video, into joint conceptual embedding spaces \cite{nayak_2021}. Our findings indicate that combining modalities  results in previously unobserved forms of representational bias, which may amplify biases existing in single-modality embedding spaces. We make our code available at \url{https://github.com/wolferobert3/visual_semantic_markedness}.

\section{Related Work}
This research draws on multiple strands of prior work on cultural markedness, CLIP and multimodal AI, and bias in AI.

\noindent\textbf{Markedness.} Visual semantic models like CLIP are designed to detect patterns in unstructured visual information, and to categorize those patterns in ways that are interpretable to humans in natural language \cite{radford2021learning}.  However, how different groups of people form categories, and what categories have utility, varies widely between communities \cite{bowker2000sorting}. While the biases of an advantaged social group may manifest explicitly in the expression of cultural stereotypes, they can also become embedded in the structure of language \cite{caliskan2020social}. One way in which this occurs is in which words and contexts are "unmarked" in language, and more natural, and which are "marked" as less natural to the speaker \cite{dressler1985predictiveness}.

The concept of "markedness" originates in structural linguistics. \citet{trubetzkoy1969principles} use the term to refer to phonological distinctions in language, wherein one in a pair of opposing phonemes has a mark, and the other lacks the mark. \citet{jakobson1972verbal} expands this to semantics, and asserts that linguistic opposites can be characterized by the presence or absence of some attribute, which is marked in a secondary form (\textit{i.e.,} a form lower in a linguistic hierarchy), but not in the dominant form (higher in the hierarchy). For example, the names of male animals (such as lion) are unmarked, while the names of female animals (lioness) are marked. \citet{greenberg1963universals} describes markedness as a consequence of the frequency with which a linguistic construction occurs. \citet{dressler1985predictiveness} describes markedness as linguistic unnaturalness, or as morphological difficulty, while \citet{givon1991markedness} finds that a marked category is cognitively more complex for the speaker.

Markedness also has sociological implications and has been studied as a sociological phenomenon. \citet{waugh1982marked} proposes a semiotic theory of markedness which suggests that marginalized social categories such as homosexuality are marked, while categories such as heterosexuality are unmarked. \citet{battistella1990markedness} generalizes this and shows that more culturally dominant terms are more broadly defined and are thus "unmarked," while "marked" terms are more narrowly defined. \citet{mayerthaler1988morphological} notes that that which is unmarked agrees with "the typical attributes of the speaker." Finally, \citet{tannen1993marked} notes that that which is culturally unmarked "carries the meaning that goes without saying," while that which is marked is unusual, and must be explained. Our work evaluates whether CLIP has learned, as \citet{battistella1990markedness} puts it, an "evaluative superstructure" of language which influences how the model embeds unstructured visual information.

\noindent\textbf{Intersectionality.} Intersectionality refers to the ways in which a person's demographic and socioeconomic characteristics combine to multiply marginalize or multiply advantage the person \cite{collins2020intersectionality}. \citet{crenshaw1989demarginalizing} introduced intersectionality to describe the multiple marginalization of black women, and the theory is used to understand the effects of marginalization along many sociocultural axes, including race, gender, sexual orientation, age, disability status, and class \cite{collins2020intersectionality}. Intersectional biases associated with members of multiply marginalized individuals may not be associated with all people who share only one of these constituent characteristics \cite{ghavami2013intersectional}. Intersectional marginalization also compounds the effects of biases associated with an individual's constituent identities \cite{crenshaw1990mapping}. Intersectional biases have been discovered in both computer vision \cite{buolamwini2018gender,steed2021image} and in language models \cite{guo2021detecting,sheng2019woman,tan2019assessing,may2019measuring}. We examine biases related to race or ethnicity, gender, and age in this research, and in some cases characterize results in terms of intersectional effects.

\noindent\textbf{CLIP and Multimodal AI.} CLIP is the first multimodal AI model to learn "transferable" visual representations, meaning that it performs at near state-of-the-art for image ranking, retrieval, and classification across many computer vision evaluation datasets, without explicitly training on the training data for those datasets. Before CLIP, the state-of-the-art for zero-shot image classification on ImageNet \cite{deng2009imagenet} was 11.6\% accuracy; CLIP improved this to 76.5\% \cite{radford2021learning}, realizing a significant advance for semi-supervised computer vision. CLIP jointly pretrains a language model (a smaller version of the GPT-2 architecture \cite{radford2019language}) and a computer vision model (either a ResNet \cite{he2016deep} or a Vision Transformer \cite{dosovitskiy2020image}), and projects the representations formed by each model independently into a joint language-and-vision embedding space. The model's training objective is to maximize the cosine similarity between a projected image and its projected natural language caption, while minimizing the similarity between the image and all of the other projected captions in the batch \cite{radford2021learning}, a training objective referred to as contrastive learning \cite{radford2021learning,tian2019contrastive}. In addition to learning transferable visual features, CLIP has also been shown to learn highly semantic word and sentence representations which set or match state-of-the-art on some intrinsic evaluations \cite{wolfe2022contrastive}. We report results on the CLIP-ViT-Base-Patch32 model, which was downloaded more than one million times in the month prior to April 26, 2022 from the Transformers library, and constituted more than 98\% of CLIP downloads from the library during that month \cite{wolf-etal-2020-transformers}.

Visual semantic embedding spaces have their origin in the research of \citet{socher2013zero}, who designed an approach for zero-shot transfer between image and language representations. \citet{frome2013devise} subsequently embed images and text in a joint embedding space in DeViSe, allowing limited generalization to unseen classes. CLIP builds on recent work in multimodal transformer models, including that of \citet{lu2019vilbert}, who introduce VilBERT, a multimodal model which extended the BERT masked language model of \citet{devlin-etal-2019-bert} to train joint "visiolinguistic" representations on the Conceptual Captions dataset \cite{sharma2018conceptual} using co-attentional transformer layers. \citet{li2019visualbert} introduce VisualBERT, a multimodal model which implicitly aligns language and images using the attention weights of a transformer and which is capable of grounding elements of language to specific regions of an image. More recently, \citet{zhang2020contrastive} employ contrastive learning with an image encoder (ResNet 50) and a contextualizing language model (BERT) to train a medical image classifier known as ConVIRT, and \citet{jia2021scaling} employ a similar design with ALIGN, an image classifier trained on contrastive loss (normalized softmax) between a BERT language model and an EfficientNet-L2 \cite{xie2020self}.  \citet{wang2021simvlm} introduce SimVLM, a multimodal encoder-decoder transformer trained only on language modeling loss, rather than contrastive loss. \citet{tiwary_2021} introduce Turing Bletchley, a multimodal model trained on contrastive loss and capable of performing language-vision tasks in 94 languages. Most recently, \citet{mu2021slip} introduce SLIP, a model which adds view-based  self-supervision of images to the contrastive learning training objective.

\noindent\textbf{Bias in AI.} Unsupervised and semi-supervised pretraining is known to encode social biases inherent in the training data \cite{caliskan2017semantics,bender2021dangers}. While CLIP is multimodal in that it combines image and text representations into a joint embedding space, the transformers used to form this space are a language model and a computer vision model \cite{radford2021learning}. Accordingly, we review biases observed in computer vision, language models, and multimodal AI. 

\noindent\textbf{\textit{Bias in Computer Vision.}} \citet{buolamwini2018gender} show that three state-of-the-art AI facial recognition systems fail to detect the faces of women with darker skin, and that common facial recognition benchmark datasets contain primarily images of people with lighter skin, where skin tone is defined based on the Fitzpatrick dermatological skin type classification system \cite{fitzpatrick1988validity}. \citet{wilson2019predictive} show that state-of-the-art object detection systems exhibit better performance for people with lighter skin than for people with darker skin. More recently, \citet{kim2021age} find that commercial emotion recognition systems detect emotion least accurately in older adults,\footnote{There are fundamental concerns with the use of facial recognition and analysis tools, which may be used for purposes of surveillance and social control \cite{scheuerman2021auto,leibold2020surveillance,guetta2021dodging}. However, the failure of such technologies to perform according to their intended use for underrepresented social groups reflects a bias of underrepresentation in the training data.} while \citet{park2021understanding} examine 92 face datasets used to train facial analysis systems, and show that less than half include at least one image of a person older than 65, and that only one dataset includes an image of a person older than 85. \citet{steed2021image} show that Image GPT \cite{chen2020generative} generates images which reflect harmful human social biases, such as sexualized pictures of women, and that it associates members of minority racial groups with weapons. \citet{wang2019balanced} find that even balancing the training dataset based on gender does not necessarily prevent deep learning models from learning gender biases. 

\noindent\textbf{\textit{Bias in Language Models.}} \citet{sheng2019woman} show that the text output of language models such as GPT-2 \cite{radford2019language} demonstrates low "regard" for women and sexual minorities. In developing a benchmark to quantify both bias and language modeling performance based on model output, \citet{nadeem2020stereoset} find that larger language models (\textit{i.e.}, those with more parameters) exhibit better language performance and prefer stereotypical human biases more than do smaller models. \citet{guo2021detecting} adapt the Word Embedding Association Test (WEAT) of \citet{caliskan2017semantics} to contextualized word embeddings by treating contextualization as a random effect. They find that word embeddings formed by language models such as BERT \cite{devlin-etal-2019-bert} encode a range of social biases, including biases related to gender and race, and that the magnitude of intersectional bias is greater than for bias based solely on gender or race. \citet{may2019measuring} measure bias in sentence vectors formed by language models and find evidence of sentence-level racial and gender biases. Recent research examines biases of representational similarity in language models. \citet{wolfe2021low} find that underrepresentation in the text training corpora of language models results in contextualized word embeddings which are more self-similar across identical contexts, yet undergo more change in the model, suggesting that language models generalize poorly to less frequently observed groups, and overfit to pretraining contexts. \citet{dodge2021documenting} show that blocklist filtering used in the construction of the C4 language modeling dataset removes information authored by and about people belonging to minority populations.

\noindent\textbf{\textit{Bias in CLIP.}} \citet{radford2021learning} and \citet{agarwal2021evaluating} find that CLIP disproportionately associates text descriptions related to appearance with images of Female individuals; that images of Black individuals are the most likely to be misclassified as animals; and that images of individuals under 20 years old are most likely to be classified into crime-related categories \cite{radford2021learning}. \citet{goh2021multimodal} identify multimodal neurons in CLIP which associate stereotypes (such as immigration) with people or regions (such as Latin America). \citet{wang2021gender} use CLIP as an image retrieval system, and find evidence of pervasive gender imbalances in retrieved images. \citet{birhane2021multimodal} show that LAION-400m \cite{schuhmann_2021}, a dataset of more than 400 million image-text pairs designed to be similar to the one on which CLIP trains, contains pornographic, misogynistic, and stereotypical images and accompanying text captions. It is unclear to what extent CLIP's training data (which has not been released to the public) was filtered for such content.

\section{Data}\label{sec:data}
In analyzing the social impact of CLIP, \citet{radford2021learning} use the FairFace dataset of \citet{karkkainen2021fairface}, designed to address the lack of racial diversity in computer vision datasets by creating nearly balanced classes for seven racial or ethnic groups. In order to compare more directly with the results reported by \citet{radford2019language}, and because FairFace is among the largest relatively balanced datasets of human images, we use the 86,744 images in the FairFace training dataset. The FairFace dataset labels images according to seven races or ethnicities, two genders, and nine age ranges:

\begin{itemize}
    \item Race or Ethnicity: Black, White, Latino or Hispanic, East Asian, Southeast Asian, Indian\footnote{In the FairFace dataset, "Indian" refers not to people of Indigenous American ancestry, but to people whose ancestry originates in India.}, Middle Eastern
    \item Gender: Female, Male
    \item Age: 0-2, 3-9, 10-19, 20-29, 30-39, 40-49, 50-59, 60-69, more than 70
\end{itemize}

For our experiments, we balance the dataset on the attribute or attributes under consideration. For example, when examining gender bias, images of Male individuals (the more frequently occurring category in FairFace) are removed without replacement until we have the same number of Male images and Female images. When examining intersectional bias operating over race, gender, and age, we obtain the number of images associated with the least frequent intersectional group and randomly remove images from every other group until they contain the same number. To mitigate the effects of downsampling, we repeat this process $10,000$ times and report the mean.

We do not endorse the categorical or binary division of human beings based on race, gender, or age. Nevertheless, we recognize that the categories defined by FairFace are broadly operational in our current social context, and thus we operationalize them here to undertake research related to the biases which attend them. Because FairFace labels are assigned to images by Amazon Mechanical Turkers, the FairFace dataset reflects how images of humans are perceived by other potentially biased humans, and does not necessarily reflect self-identification.

\noindent\textbf{CLIP Training Data.} The patterns learned by unsupervised and semi-supervised machine learning models are dependent on the language and society which produces the training data. \citet{radford2021learning} train CLIP on the WebImageText corpus (WIT), a web scrape composed of $400$ million images and associated captions. \citet{radford2021learning} produce the query list using every word which occurs at least 100 times in English Wikipedia, plus bigrams from Wikipedia with high pointwise mutual information, the names of Wikipedia articles, and all WordNet synsets \cite{radford2021learning}. The frequency-based heuristic for inclusion in WIT suggests that biases related to underrepresentation may occur in CLIP.

\section{Approach and Experiments}
We describe an approach for measuring self-similarity and testing whether CLIP exhibits cultural markedness.

\noindent\textbf{Representational Similarity.} We obtain a projected image embedding from every image in the Fairface dataset, and measure the self-similarity (\textit{i.e.,} the mean pairwise cosine similarity) of projected image embeddings for each race or ethnicity, gender, and age range, as well as for the intersectional groups in the dataset. Specifically, for a set of images, the self-similarity $s$ for a group of images $G$ consisting of $n$ images $i \in G$ is given by: 

\begin{equation}
    s(G) = \frac{1}{n^2 - n} \sum_{i} \sum_{j \neq i}  cos(\vec{i_i}, \vec{i_j})
\label{self_sim_equation}
\end{equation}

\noindent $cos$ in Equation \ref{self_sim_equation} refers to cosine similarity, or the angular similarity of two vectors after normalization to unit length. Note that image representations projected to the visual semantic space in CLIP are multimodal, meaning that they can be meaningfully compared not only to other images, but to text encoded by the model as well. Prior work indicates that, as embedded images reach the top layers of the CLIP image encoder before projection, they acquire abstract conceptual features, which activate in response to textual, photographic, and symbolic representations of the depicted concept and other similar concepts \cite{goh2021multimodal}. For example, \citet{goh2021multimodal} find that the same multimodal neuron activates in the upper layers of the CLIP image encoder for cartoon depictions of Spider-Man, text of the word "spider," and photographs of humans dressed as Spider-Man. What this suggests for the present research is that, in a multimodal embedding space, self-similarity is a meaningful measurement for evaluating the extent to which a group is culturally marked by a shared conceptual characteristic: if, for example, people over the age of 70 are marked by CLIP based on age, we would expect that a group of images of people over 70 embedded in a visual semantic embedding space would be more similar to each other (by virtue of the shared, marked characteristic) than would a group of images of people in their 30s. Equation \ref{self_sim_equation} measurement has been used previously to measure bias in language models such as BERT \cite{devlin-etal-2019-bert} and T5 \cite{raffel2020exploring}, which were shown to embed names statistically more associated with underrepresented racial and gender groups such that they are more self-similar across contexts, reflecting that the range of representations is more narrowly defined \cite{wolfe2021low}. To our knowledge, this is the first application of the measurement to study bias in multimodal AI.

\noindent\textbf{Markedness According to Race or Ethnicity, Gender, and Age.}  For each image in the FairFace dataset, we obtain a projected image embedding. Then, we obtain projected text embeddings corresponding to every race or ethnicity label, every gender label, and every age label in the FairFace dataset, as described in Section \ref{sec:data}. \citet{radford2021learning} find that, because single-word labels are uncommon in CLIP's dataset, the model performs better in the zero-shot setting when used with the prompt "a photo of a [image class]." To that end, we use the prompt "a photo of a [race or ethnicity] person" for race or ethnicity labels; "a photo of a [gender] person" for gender labels; and "a photo of a person between [age range minimum] and [age range maximum] years old" for age labels. For the "More than 70" age range, we use the prompt "a photo of a person more than 70 years old." Finally, we obtain a projected text embedding for the text label "a photo of a person," with no information related to race or ethnicity, gender, or age.

The cosine similarity is then computed between each projected text embedding and the projected image embedding. Cosine similarity is a direct proxy to the probability that an image is accurately associated with a text label in CLIP, and is used by the model to rank, retrieve, and classify images \cite{radford2021learning,wang2021gender}. For each image, we observe whether the cosine similarity of the projected image embedding is higher with the "a photo of a person" label than with any of the race or ethnicity labels; than with any of the gender labels; and than with any of the age range labels. Our research is not concerned with whether CLIP has classified in accordance with the FairFace label for a set of socially defined categories. Rather, it measures whether the model associates an image of a person with \textit{any} of the labels corresponding to a socially defined category, or if the image is left unmarked. Results report the percentage of the time a social group is unmarked (\textit{i.e.,} the "a photo of a person" label is ranked highest by CLIP according to cosine similarity) for a given social category. For example, results will show that individuals who are White and Male and between the ages of 30 and 39 remain unmarked based on race (the "a photo of a person" label is preferred) 46.6\% of the time. We report results for each race or ethnicity, gender, and age range in the FairFace dataset, and examine results from an intersectional lens to observe whether, for example, the age of a person has an effect on whether they are marked with a gender label. 

\section{Results}

The evidence indicates that CLIP embeds images of women, of relatively younger and older people, and of people belonging to racial and ethnic groups other than White such that they are more self-similar. CLIP consistently prefers to mark the race or ethnicity of racial groups other than White, the gender of women, and age of relatively young or old people, while leaving images of people who are White, Male, and middle-aged unmarked.

\noindent\textbf{Representational Uniformity.} The evidence reveals biases of representational uniformity which differ based on race or ethnicity, gender, and age. Table \ref{fairface_race_self_sim} indicates that the most variation ($s=0.573$) exists in the multimodal image embeddings of White individuals, while embeddings of Indian, East Asian, and Southeast Asian individuals are the most self-similar to each other, with $s > 0.620$. Table \ref{fairface_gender_self_sim} indicates that embedded images of Male individuals are less self-similar ($s=0.566$) than embedded images of Female individuals ($s=0.592$). Table \ref{fairface_age_self_sim} describes differences in self-similarity based on age. Most notably, images of individuals between the ages of 3 and 9 are much more self-similar ($s=0.669$) than those of individuals between 20 and 29, 30 and 39, 40 and 49, and 50 and 59 ($s \in [0.573,0.575]$).  A similar phenomenon manifests for individuals between the ages of 10 and 19, for which self-similarity is higher ($s=0.611$) than for images of individuals between 20 and 59, despite the significant physical variation of people aged 10-19.

\begin{table*}[htbp]
{
\centering
\begin{tabular}
{|l||r|r|r|r|r|r|r|}
 \hline
 \multicolumn{8}{|c|}{Self-Similarity by Race or Ethnicity} \\
\hline
 Metric & White & Latino/a & Middle Eastern & Black & Southeast Asian & East Asian & Indian\\
 \hline
 Self-Similarity & \cellcolor{gray!10}0.573 & \cellcolor{gray!18}0.589 & \cellcolor{gray!18}0.594 & \cellcolor{gray!22}0.604 & \cellcolor{gray!30}0.623 & \cellcolor{gray!30}0.623 & \cellcolor{gray!34}0.626 \\
 \hline
\end{tabular}
}
\caption{CLIP representations of White individuals are less self-similar than representations of people of any other race or ethnicity, indicating that CLIP directs less attention to the race of White individuals when encoding multimodal representations, as it does not cluster them as closely together based on the shared characteristic of race.}
\label{fairface_race_self_sim}
\Description{Table of Self-Similarity by Race or Ethnicity}
\end{table*}

\begin{table*}[htbp]
{
\centering
\begin{tabular}
{|l||r|r|r|r|r|r|r|r|r|}
 \hline
 \multicolumn{10}{|c|}{Self-Similarity by Age Range} \\
\hline
 Metric & 0-2 & 3-9 & 10-19 & 20-29 & 30-39 & 40-49 & 50-59 & 60-69 & 70+\\
\hline
 Self-Similarity & \cellcolor{gray!58}0.692 & \cellcolor{gray!50}0.669 & \cellcolor{gray!26}0.611 & \cellcolor{gray!14}0.575 & \cellcolor{gray!10}0.574 & \cellcolor{gray!10}0.573 & \cellcolor{gray!10}0.574 & \cellcolor{gray!14}0.583 & \cellcolor{gray!30}0.623 \\
 \hline
\end{tabular}
}
\caption{CLIP representations of individuals under 20 or over 70 are more self-similar than representations of people of any other age, indicating that CLIP directs more attention to the age of relatively younger and older individuals.}
\label{fairface_age_self_sim}
\Description{Table of Self-Similarity by Age Range}
\end{table*}

\begin{table*}[htbp]
{
\centering
\begin{tabular}{|l|r|r|}
\hline
 \multicolumn{3}{|c|}{Self-Similarity by Gender} \\
 \hline
  Metric &  Male & Female  \\
\hline
  Self-Similarity &  \cellcolor{gray!10}0.566 & \cellcolor{gray!18}0.592 \\
  \hline
\end{tabular}
}
\caption{CLIP representations of Female individuals are more self-similar than representations of Male individuals, indicating that the model directs more attention to gender in forming representations of women.}
\label{fairface_gender_self_sim}
\Description{Table of Self-Similarity by Gender}
\end{table*}

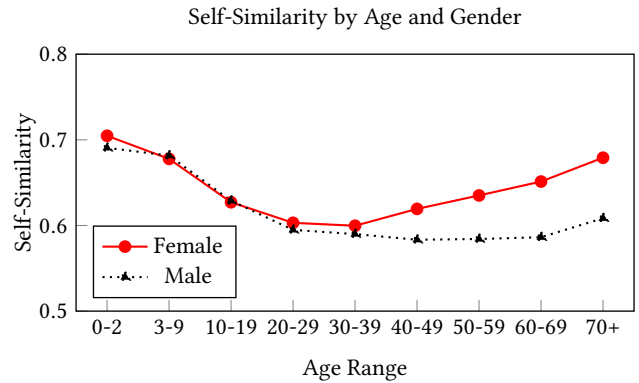
\begin{figure}[htbp]
\begin{tikzpicture}
\begin{axis} [
    height=5cm,
    width=9cm,
    line width = .5pt,
    ymin = .5, 
    ymax = .8,
    xmin=.5,
    xmax=9.5,
    ylabel=Self-Similarity,
    ylabel shift=-5pt,
    ylabel near ticks,
    xtick = {1,2,3,4,5,6,7,8,9},
    xticklabels = {0-2,3-9,10-19,20-29,30-39,40-49,50-59,60-69,70+},
    xtick pos=left,
    ytick pos = left,
    title=Self-Similarity by Age and Gender,
    xlabel= {Age Range},
    legend style={at={(.03,.05)},anchor=south west}
]
\addplot[thick,solid,mark=*,color=red] coordinates {(1,0.7047204976276811) (2,0.6777873806325028) (3,0.627119497764504) (4,0.6030972370698332) (5,0.5996736154457165) (6,0.6193575675914582) (7,0.6350797853985569) (8,0.6513004909262443) (9,0.6791195171721207)};
\addplot[thick,dotted,mark=triangle*,color=black] coordinates {(1,0.6906652853092737) (2,0.6816359804195957) (3,0.6288959447872687) (4,0.5948632604105685) (5,0.5899478109623553) (6,0.5833336956364031) (7,0.5842125023304849) (8,0.5863068557710769) (9,0.6083971851360463)};
\legend {Female, Male};

\end{axis}
\end{tikzpicture}
\caption{At each increase in age range beyond 20-29, the disparity in self-similarity between Female individuals and Male individuals increases, indicating that age exacerbates differences based on gender  in CLIP representations.}
\label{gender_by_age_range}
\Description{Figure depicting increased gender disparity in self-similarity as age increases.}
\end{figure}

Examining the data from an intersectional lens reveals additional disparities. Figure \ref{gender_by_age_range} indicates that images of Female individuals are more self-similar when compared within the same age range against Male individuals, a difference which is first evident for the $20-29$ age range. With each increase in age range, the difference in self-similarity between Female individuals and Male individuals also increases: from 0.036 in the 40-49 age range ($s=0.619$ vs. $s=0.583$), to 0.051 in the 50-59 age range ($s=0.635$ vs. $s=0.584$), to 0.065 in the 60-69 age range ($s=0.651$ vs. $s=0.586$), to 0.071 in the 70-79 age range ($s=0.679$ vs. $s=0.608$). The data indicate a gender bias which affects representations of Female individuals as age increases.

\begin{figure}[htbp]
\begin{tikzpicture}
\begin{axis} [
    height=5cm,
    width=9cm,
    line width = .5pt,
    ymin = .5, 
    ymax = .8,
    xmin=.5,
    xmax=9.5,
    ylabel=Self-Similarity,
    ylabel shift=-5pt,
    ylabel near ticks,
    xtick = {1,2,3,4,5,6,7,8,9},
    xticklabels = {0-2,3-9,10-19,20-29,30-39,40-49,50-59,60-69,70+},
    xtick pos=left,
    ytick pos = left,
    title=Self-Similarity by Age and Race or Ethnicity,
    xlabel= {Age Range},
    legend style={at={(0.45,0.57)},anchor=south west,nodes={scale=0.6, transform shape}},
         title style={yshift=2mm, align=center}
]

\addplot[thick,solid,mark=*,color=red] coordinates {(1,0.726509177935345) (2,0.7213586501902269) (3,0.6585771851963198) (4,0.6363246111868316) (5,0.6292116161863566) (6,0.644454213690861) (7,0.6474633325353071) (8,0.6656806642706935) (9,0.7128647188212813)};
\addplot[thick,solid,mark=x,color=violet] coordinates {(1,0.7284054232898817) (2,0.6809621419473627) (3,0.6181740253791896) (4,0.5887704440137298) (5,0.5926512091405834) (6,0.5975864126022893) (7,0.5911525166956353) (8,0.5966601783678074) (9,0.6280633508807806)};
\addplot[thick,dashed,mark=square*,color=orange] coordinates {(1,0.7089316557130839) (2,0.6956565870652229) (3,0.6464632358046687) (4,0.6074751776085979) (5,0.6048467544464962) (6,0.607464855782074) (7,0.6234238858162231) (8,0.631767658738614) (9,0.6708574742700968)};
\addplot[thick,dashdotted,mark=+,color=blue] coordinates {(1,0.7367972462808209) (2,0.7277331053345771) (3,0.6996356661696028) (4,0.6357153593696786) (5,0.6422876218010274) (6,0.637402299959708) (7,0.6417073560034685) (8,0.6581345061993923) (9,0.7050332896395153)};
\addplot[thick,dotted,mark=diamond*,color=green] coordinates {(1,0.7431708940816898) (2,0.7402820433038241) (3,0.6697304116159892) (4,0.6043214719982823) (5,0.5963618751884981) (6,0.6043706352946111) (7,0.6102527862869099) (8,0.6372059517277501) (9,0.6623083905511102)};
\addplot[thick,dotted,mark=triangle*,color=black] coordinates {(1,0.736194711122647) (2,0.714683339015205) (3,0.646676302176758) (4,0.6399429500652869) (5,0.6425539181426791) (6,0.6442799600595691) (7,0.6525010613191573) (8,0.6571731183966993) (9,0.6764486827010853)};
\addplot[thick,dashed,mark=otimes*,color=yellow] coordinates {(1,0.7411997619797149) (2,0.687371634001906) (3,0.6350373735792559) (4,0.6182896958344979) (5,0.6186509014812955) (6,0.6050994922358341) (7,0.6061679180043971) (8,0.6125437646752165) (9,0.6361839490779265)};
\legend {Southeast Asian, White, Latino or Hispanic, Indian, Black, East Asian, Middle Eastern};

\end{axis}
\end{tikzpicture}
\caption{At each increase in age range beyond 40-49, the disparity in self-similarity between White individuals and people belonging to the most self-similar race or ethnicity (Southeast Asian) also increases, indicating that age exacerbates differences based on race or ethnicity in CLIP representations.}
\label{race_age_figure}
\Description{Figure depicting increased race or ethnicity disparity in self-similarity as age increases.}
\end{figure}
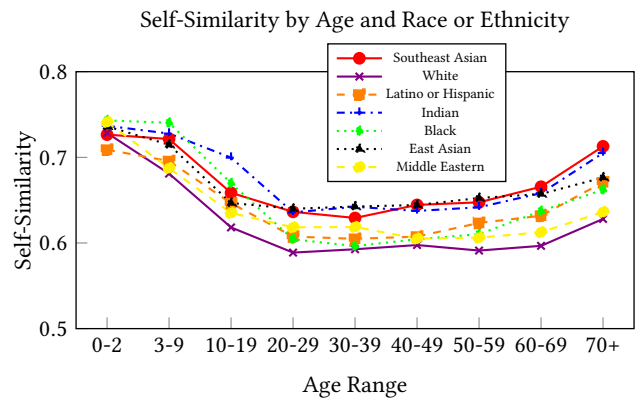

Figure \ref{race_age_figure} shows that embedded images of White individuals  are the least self-similar at every age except for  0-2. As with the gender bias, increases in age exacerbate already existing differences based on race. For individuals between 40 and 49, the difference between the most self-similar race or ethnicity (Southeast Asian, $s=0.644$) and the least self-similar race or ethnicity (White, $s=0.598$) is 0.046; this difference increases with each increase in age, such that self-similarity for individuals who are White and more than 70 is 0.628, while self-similarity for individuals who are Southeast Asian and more than 70 is 0.713, a difference of 0.085.

We face a challenge in comprehensibly describing the intersectional effects of self-similarity given the significant diversity in race and ethnicity, gender, and age in the dataset we use. A partial description of these results is given in Table \ref{intersectional_most_similar}, which reports the self-similarity of projected image embeddings for the most self-similar and the least self-similar social groups. In addition to these tables, we visualize the difference in self-similarity by age for individuals who are White and Male and for individuals who are Black and Female. The selection of these groups is not arbitrary. \citet{crenshaw1989demarginalizing} introduce intersectionality by describing the overlapping axes of discrimination faced by Black women, while White men have enjoyed economic, political, and cultural advantages based on race and gender. If an intersectional effect exists which compounds the effects noted for race, gender, and age, we would expect to be able to observe this via a comparison of these two social groups.

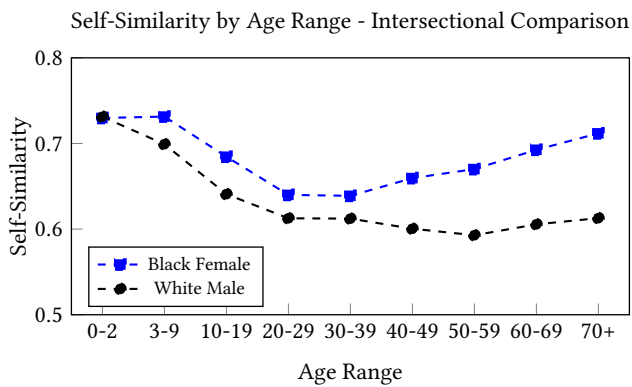
\begin{figure}[htbp]
\begin{tikzpicture}
\begin{axis} [
    height=5cm,
    width=9cm,
    line width = .5pt,
    ymin = .5, 
    ymax = .8,
    xmin=.5,
    xmax=9.5,
    ylabel=Self-Similarity,
    ylabel shift=-5pt,
    ylabel near ticks,
    xtick = {1,2,3,4,5,6,7,8,9},
    xticklabels = {0-2,3-9,10-19,20-29,30-39,40-49,50-59,60-69,70+},
    xtick pos=left,
    ytick pos = left,
    title=Self-Similarity by Age Range - Intersectional Comparison,
    xlabel= {Age Range},
    legend style={at={(.03,.03)},anchor=south west,nodes={scale=0.8, transform shape}}
]
\addplot[thick,dashed,mark=square*,color=blue] coordinates {(1,0.7297218356911839) (2,0.7314027009848392) (3,0.6845345598679948) (4,0.6399686703425392) (5,0.6388680341948526) (6,0.6595439929013457) (7,0.6700626839889924) (8,0.6926178080369502) (9,0.7117994411444923)};

\addplot[thick,dashed,mark=otimes*,color=black] coordinates {(1,0.7312194317539897) (2,0.6991331307072572) (3,0.6408138269507349) (4,0.6126322639288833) (5,0.6122332238620596) (6,0.6004596184672714) (7,0.5928440880139574) (8,0.60557230231904) (9,0.6127989684561103)};
\legend {Black Female, White Male};

\end{axis}
\end{tikzpicture}
\caption{At every age range after 0-2, CLIP representations of individuals who are Black and Female are more self-similar than representations of individuals who are White and Male, and the disparity between the groups is greater than the disparities between Male individuals and Female depicted in Figure \ref{gender_by_age_range}.}
\label{wm_bf_intersectional_figure}
\Description{Figure depicting increased intersectional (gender and race) disparity in self-similarity as age increases.}
\end{figure}

Figure \ref{wm_bf_intersectional_figure} demonstrates that disparities in self-similarity are exacerbated in the intersection of race, gender, and age. The same divergence in self-similarity exists as was observed for gender in Figure \ref{gender_by_age_range}. However, the magnitude of the difference is greater at every age range, including much greater differences for individuals between the ages of 3 and 39, which are relatively small when observed solely for gender. Concretely, the self-similarity of images of individuals who are White and Male and 70 or older is 0.613, while the self-similarity of images of individuals who are Black and Female and 70 or older is 0.712, a difference of 0.099, higher than the difference based on gender for individuals 70 and older (.071) and the largest difference based on race for individuals 70 and older (.085).

\begin{table*}[htbp]
{
\centering
\begin{tabular}
{|l|c||l|c|}
 \hline
 \multicolumn{4}{|c|}{Self-Similarity by Social Group} \\
\hline
 \multicolumn{2}{|c||}{Most Self-Similar Social Groups} &  \multicolumn{2}{|c|}{Least Self-Similar Social Groups} \\
 \hline
  \multicolumn{1}{|c|}{Social Group} & Self-Similarity & \multicolumn{1}{|c|}{Social Group} & Self-Similarity\\
\hline
Black Male 0-2 & \cellcolor{gray!90}0.767 & White Male 70+ & \cellcolor{gray!26}0.613\\
Black Male 3-9  & \cellcolor{gray!86}0.759 & White Male 20-29 & \cellcolor{gray!26}0.613\\
Southeast Asian Female 70+ & \cellcolor{gray!86}0.757 & Latino or Hispanic Male 30-39 &  \cellcolor{gray!26}0.612\\
East Asian Female 0-2 & \cellcolor{gray!82}0.751 & White Male 30-39 & \cellcolor{gray!26}0.612\\
Indian Female 0-2 & \cellcolor{gray!82}0.751 & Black Male 50-59 & \cellcolor{gray!26}0.608\\
Middle Eastern Male 0-2 & \cellcolor{gray!82}0.748 & White Male 60-69 & \cellcolor{gray!26}0.606\\
Southeast Asian Female 0-2 & \cellcolor{gray!82}0.747 & Black Male 30-39 & \cellcolor{gray!22}0.603\\
Indian Female 70+ & \cellcolor{gray!82}0.746 &  White Male 40-49 & \cellcolor{gray!22}0.600\\
White Female 0-2 & \cellcolor{gray!78}0.744 & Black Male 20-29 & \cellcolor{gray!22}0.600\\
Indian Male 3-9 & \cellcolor{gray!78}0.742 & White Male 50-59 &       \cellcolor{gray!18}0.593\\
 \hline
\end{tabular}
}
\caption{The most self-similar social groups reflect individuals younger than 10 or older than 70, and predominantly Female individuals. The least self-similar social groups all reflect Male individuals, and predominantly White individuals and individuals between the ages of 20 and 59.}
\label{intersectional_most_similar}
\Description{Table displaying the most self-similar and least self-similar social groups.}
\end{table*}

Table \ref{intersectional_most_similar} indicates that the ten least self-similar social groups all reflect Male individuals; that they reflect only three of the seven races or ethnicities in the dataset (White, Black, and Latino or Hispanic); and that eight of the ten groups come from age ranges between 20 and 59 years old, with no age ranges under the age of 20 represented. Six of the ten social groups reflect individuals who are White and Male, including the only two groups of individuals over the age of 60. The most self-similar social groups all reflect people who are younger than age of 10 or older than the age of 70. Unlike the ten least self-similar groups, six of the ten most self-similar groups reflect Female individuals, and only one of the groups reflects White individuals.

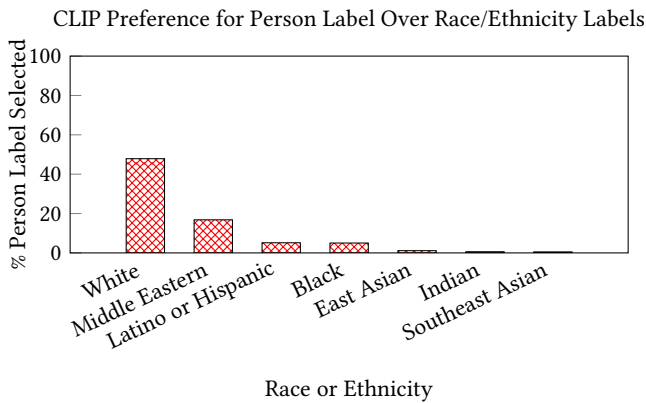
\begin{figure}[htbp]
\begin{tikzpicture}
\begin{axis} [
    height=4.2cm,
    width=9cm,
    ybar = .05cm,
    bar width = 14.5pt,
    ymin = 0, 
    ymax = 100,
    ylabel=\% Person Label Selected,
    ylabel shift=-5pt,
    ylabel near ticks,
    xtick = {1,2,3,4,5,6,7},
    xtick style={draw=none},
    ytick pos = left,
    xticklabels = {White, Middle Eastern, Latino or Hispanic, Black, East Asian, Indian, Southeast Asian},
    xticklabel style={rotate=25,anchor=east},
    title=CLIP Preference for Person Label Over Race/Ethnicity Labels,
    x label style={at={(axis description cs:0.5,-0.4)},anchor=north},
    xlabel= {Race or Ethnicity},
    enlarge x limits={abs=1cm}
]

\addplot [pattern=crosshatch,pattern color = red] coordinates {(1,47.866319444444444) (2,16.80772569444444) (3,5.152343749999999) (4,5.018880208333334) (5,1.1378038194444447) (6,0.5865885416666668) (7,0.5062934027777778)};

\end{axis}
\end{tikzpicture}
\caption{CLIP prefers a label with race or ethnicity omitted less than 6\% of the time for five of seven racial or ethnic groups, but prefers a label with no information about race or ethnicity 47.9\% of the time for White individuals.}
\label{preference_race}
\Description{Figure depicting CLIP's preference for a person label over a race or ethnicity label, broken down by race or ethnicity.}
\end{figure}

\begin{figure}[htbp]
\begin{tikzpicture}
\begin{axis} [
    height=4.2cm,
    width=7cm,
    ybar = .05cm,
    bar width = 30.5pt,
    ymin = 0, 
    ymax = 100,
    ylabel=\% Person Label Selected,
    ylabel shift=-5pt,
    ylabel near ticks,
    xtick = {1,2},
    xtick style={draw=none},
    ytick pos = left,
    xticklabels = {Male,Female},
    xlabel= Gender,
    title=CLIP Preference for Person Label Over Gender Labels,
    enlarge x limits=1
]

\addplot [pattern=horizontal lines,pattern color = green] coordinates {(1,26.66202954021296) (2,15.238726139653563)};

\end{axis}
\end{tikzpicture}
\caption{CLIP prefers the "person" label over labels with gender information 26.7\% of the time for Male individuals, compared to 15.2\% of the time for Female individuals, indicating that CLIP marks the gender of women at higher rates than men.}
\label{preference_gender}
\Description{Figure depicting CLIP's preference for a person label over a gender label, broken down by gender.}
\end{figure}
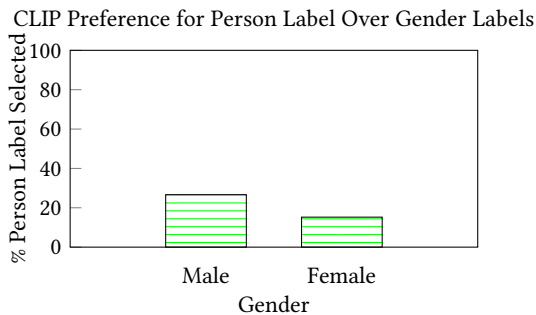

\noindent\textbf{Markedness Based on Race or Ethnicity, Gender, and Age.} As shown in Figure \ref{preference_race}, CLIP ranks the “a photo of a person” label highest 47.9\% of the time for images of White individuals. For images of individuals of every other race or ethnicity in FairFace, CLIP ranks “a photo of a person” highest less than 16.8\% of the time, and less than 1.5\% of the time for East Asian, Indian, or Southeast Asian individuals. CLIP ranks “a photo of a person” highest 26.7\% of the time for images of Male individuals, but 15.2\% of the time for images of Female individuals, as shown in Figure \ref{preference_gender}, reflecting a preference to mark the gender of Female individuals.

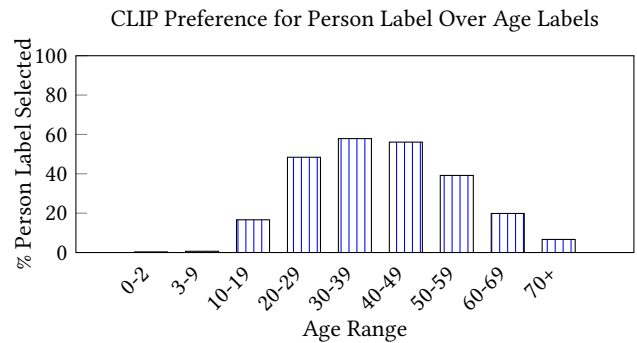
\begin{figure}[htbp]
\begin{tikzpicture}
\begin{axis} [
    height=4.2cm,
    width=9cm,
    ybar = .05cm,
    bar width = 12.5pt,
    ymin = 0, 
    ymax = 100,
    ylabel=\% Person Label Selected,
    ylabel shift=-5pt,
    ylabel near ticks,
    xtick = {1,2,3,4,5,6,7,8,9},
    xtick style={draw=none},
    ytick pos = left,
    xticklabels = {0-2,3-9,10-19,20-29,30-39,40-49,50-59,60-69,70+},
    xticklabel style={rotate=45,anchor=east},
    x label style={at={(axis description cs:0.5,-0.1)},anchor=north},
    title=CLIP Preference for Person Label Over Age Labels,
    xlabel= {Age Range},
    enlarge x limits={abs=1cm}
]

\addplot [pattern=vertical lines,pattern color = blue] coordinates {(1,0.323040380047506) (2,0.63895486935866995) (3,16.63064133016627) (4,48.40855106888362) (5,57.88598574821854) (6,56.09501187648457) (7,39.16745843230405) (8,19.858669833729217) (9,6.650831353919241)};

\end{axis}
\end{tikzpicture}
\caption{CLIP prefers the "person" label over labels with age information at least 39\% of the time for individuals between 20 and 29, 30 and 39, 40 and 49, and 50 and 59, but less than 7\% of the time for individuals over 70, and less than 1\% of the time for individuals under 10.}
\label{preference_age}
\Description{Figure depicting CLIP's preference for a person label over an age label, broken down by age.}
\end{figure}

As shown in Figure \ref{preference_age}, CLIP ranks the “a photo of a person” label highest 48.4\% of the time for individuals between the ages of 20-29, 57.9\% of the time for images of individuals between the ages of 30-39, 56.1\% for images of individuals 40-49, and 39.2\% of the time for individuals 50-59. CLIP ranks the “a photo of a person” label highest less than 1\% of the time for images of individuals under 2 years old or between 3 years old and 9 years old, and 6.7\% of the time for images of individuals over 70 years old. The data suggests a bias which preferentially draws attention to the age of individuals younger than 20 or older than 59.

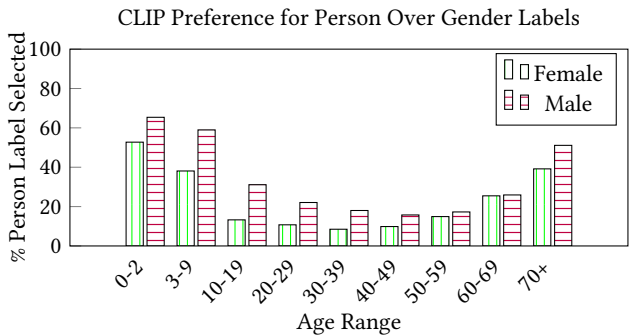
\begin{figure}[!htbp]
\begin{tikzpicture}
\begin{axis} [
    height=4.2cm,
    width=9cm,
    ybar = .05cm,
    bar width = 6.5pt,
    ymin = 0, 
    ymax = 100,
    ylabel=\% Person Label Selected,
    ylabel shift=-5pt,
    ylabel near ticks,
    xtick = {1,2,3,4,5,6,7,8,9},
    xtick style={draw=none},
    ytick pos = left,
    xticklabels = {0-2,3-9,10-19,20-29,30-39,40-49,50-59,60-69,70+},
    xticklabel style={rotate=45,anchor=east},
    title=CLIP Preference for Person Over Gender Labels,
    xlabel= {Age Range},
    x label style={at={(axis description cs:0.5,-0.1)},anchor=north},
    enlarge x limits={abs=1cm}
]

\addplot [pattern=vertical lines,pattern color = green] coordinates {(1,52.737196029776676) (2,38.09374689826303) (3,13.256997518610422) (4,10.7427047146402) (5,8.493424317617865) (6,9.857369727047146) (7,14.922828784119108) (8,25.469354838709673) (9,39.16952853598015)};

\addplot [pattern=horizontal lines,pattern color = purple] coordinates {(1,65.37238213399503) (2,58.96558312655087) (3,31.124392059553347) (4,22.076625310173696) (5,18.015508684863523) (6,15.751191066997519) (7,17.280843672456577) (8,25.919900744416875) (9,51.1166253101737)};

\legend {Female, Male};

\end{axis}
\end{tikzpicture}
\caption{Female individuals are more likely to be marked with a gender label at every age range, and are marked based on gender at earlier ages than are Male individuals.}
\label{preference_gender_by_age}
\Description{Figure depicting CLIP's preference for a person label over a gender label, broken down by gender and age.}
\end{figure}

Examination of the results based on gender and age yields additional insight. Figure \ref{preference_gender_by_age} shows that, at every age range, CLIP prefers the "a photo of a person" label at a higher rate for Male individuals than for Female individuals. For both gender groups, images of individuals under the age of 10 or over the age of 70 are the least likely to be marked based on gender. Images of Female individuals between the ages of 30 and 39 are the most likely to be marked based on gender, with a gender label ranked higher than the "person" label 91.5\% of the time. Images of Female individuals between the ages of 10 and 19, 20 and 29, 40 and 49, and 50 and 59 are marked based on gender more than 85.0\% of the time. The largest disparities between the Male and Female groups occur in the 3-9 and 10-19 age ranges. In the 3-9 age range, 59.0\% of Male individuals are most associated with the "a photo of a person" label, compared to 38.1\% of Female individuals. In the 10-19 age range, 31.1\% of Male individuals are most associated with the "a photo of a person" label, compared to 13.3\% of Female individuals, reflecting that CLIP marks the gender of Female individuals at an earlier age than it does Male individuals. Differences between Male individuals and Female individuals diminish in the 50-59 and 60-69 age ranges, before diverging again in the 70-79 age range, with 39.2\% of Female images unmarked compared to 51.1\% of Male images.

\begin{figure}[htbp]
\begin{tikzpicture}
\begin{axis} [
    height=4.2cm,
    width=9cm,
    ybar = .05cm,
    bar width = 6.5pt,
    ymin = 0, 
    ymax = 100,
    ylabel=\% Person Label Selected,
    ylabel shift=-5pt,
    ylabel near ticks,
    xtick = {1,2,3,4,5,6,7,8,9},
    xtick style={draw=none},
    ytick pos = left,
    xticklabels = {0-2,3-9,10-19,20-29,30-39,40-49,50-59,60-69,70+},
    xticklabel style={rotate=45,anchor=east},
    title=CLIP Preference for Person Over Age Labels,
    xlabel= {Age Range},
    x label style={at={(axis description cs:0.5,-0.1)},anchor=north},
    enlarge x limits={abs=1cm}
]

\addplot [pattern=vertical lines,pattern color = green] coordinates {(1,0.14439205955334986) (2,1.0064267990074443) (3,20.50439205955335) (4,49.595831265508686) (5,54.28915632754343) (6,45.057841191067) (7,25.26017369727047) (8,10.487196029776674) (9,2.050347394540943)};

\addplot [pattern=horizontal lines,pattern color = purple] coordinates {(1,0.4525806451612904) (2,0.37868486352357317) (3,12.251563275434243) (4,47.36377171215881) (5,60.95059553349876) (6,62.19367245657569) (7,47.027791563275436) (8,25.519503722084362) (9,11.662531017369728)};

\legend {Female, Male};

\end{axis}
\end{tikzpicture}
\caption{Female individuals are less likely to be marked with an age label prior to the age of 30, but are more likely to be marked with an age label after the age of 30, with the most significant disparity in the 50-59 age range.}
\label{preference_age_by_age}
\Description{Figure depicting CLIP's preference for a person label over an age label, broken down by gender and age.}
\end{figure}
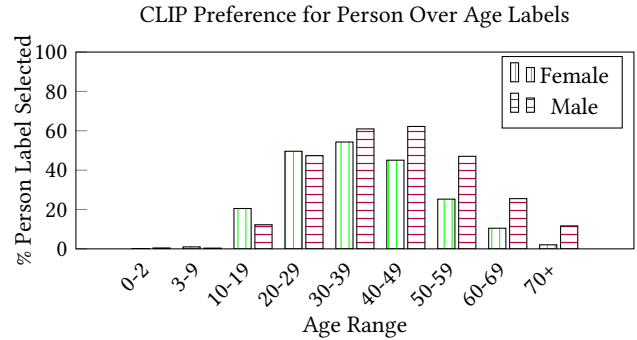

As shown in Figure \ref{preference_age_by_age}, the least likely age ranges to be marked based on gender are the most likely to be marked based on age. At least 99\% of individuals under the age of 10 are marked based on age, regardless of gender. In the 10-19 age range, CLIP ranks the "a photo of a person" label highest 20.7\% of the time for Female individuals, compared to 12.3\% of the time for Male individuals. In this age range, Female individuals are more likely to be marked based on gender, but less likely to be marked based on age, than Male individuals. Among Female individuals, CLIP ranks the "a photo of a person" label highest most frequently for people between 30 and 39, at 54.3\%; among Male individuals, CLIP ranks the "a photo of a person" label highest most frequently for people between 40 and 49, at 62.2\%. The most significant disparities between Male individuals and Female individuals occur in the 40-49 (62.2\% vs. 45.1\% unmarked) and 50-59 (47.4\% vs. 25.3\% unmarked) age ranges.

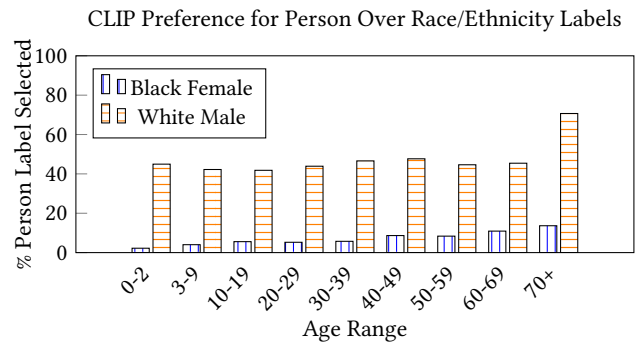
\begin{figure}[!htbp]
\begin{tikzpicture}
\begin{axis} [
    height=4.2cm,
    width=9cm,
    ybar = .05cm,
    bar width = 6.5pt,
    ymin = 0, 
    ymax = 100,
    ylabel=\% Person Label Selected,
    ylabel shift=-5pt,
    ylabel near ticks,
    xtick = {1,2,3,4,5,6,7,8,9},
    xtick style={draw=none},
    ytick pos = left,
    xticklabels = {0-2,3-9,10-19,20-29,30-39,40-49,50-59,60-69,70+},
    xticklabel style={rotate=45,anchor=east},
    x label style={at={(axis description cs:0.5,-0.1)},anchor=north},
    title=CLIP Preference for Person Over Race/Ethnicity Labels,
    xlabel= {Age Range},
    legend style={at={(0.03,0.58)},anchor=south west,nodes={scale=.95, transform shape}},
    enlarge x limits={abs=1cm}
]

\addplot [pattern=vertical lines,pattern color = blue] coordinates {(1,2.229090909090909) (2,4.0236363636363635) (3,5.568181818181818) (4,5.247272727272727) (5,5.737272727272727) (6,8.657272727272728) (7,8.367727272727272) (8,10.921363636363637) (9,13.651818181818182)};

\addplot [pattern=horizontal lines,pattern color = orange] coordinates {(1,44.94681818181817) (2,42.22772727272728) (3,41.81727272727272) (4,43.89590909090909) (5,46.6) (6,47.66954545454546) (7,44.632272727272735) (8,45.42272727272728) (9,70.68227272727273)};

\legend {Black Female, White Male};

\end{axis}
\end{tikzpicture}
\caption{In every age range, CLIP prefers a label with race or ethnicity omitted for people who are White and Male at higher rates than for individuals who are Black and Female. As age increases, the probability of individuals who are Black and Female receiving a label with race or ethnicity omitted also increases.}
\label{intersection_race_by_age}
\Description{Figure depicting CLIP's preference for a person label over a race label, broken down by race, gender, and age.}
\end{figure}

Finally, we report results based on race or ethnicity, gender, and age for individuals who are White and Male, and for individuals who are Black and Female. As shown in Figure \ref{intersection_race_by_age}, CLIP ranks the "a photo of a person" label highest at least 40\% of the time in every age range for individuals who are White and Male, and less than 15\% of the time in all age ranges for individuals who are Black and Female. A pattern emerges for individuals who are Black and Female: as age increases, the probability of CLIP preferring the "a photo of a person" label over a label denoting race or ethnicity also increases, from 2.2\% at 0-2, to 5.7\% at 30-39, to 13.7\% at more than 70. The pattern is less consistent for individuals who are White and Male, but increases significantly, from 45.4\% to 70.7\%, from the 60-69 age range to the more than 70 age range.

\begin{figure}[!htbp]
\begin{tikzpicture}
\begin{axis} [
    height=4.2cm,
    width=9cm,
    ybar = .05cm,
    bar width = 6.5pt,
    ymin = 0, 
    ymax = 100,
    ylabel=\% Person Label Selected,
    ylabel shift=-5pt,
    ylabel near ticks,
    xtick = {1,2,3,4,5,6,7,8,9},
    xtick style={draw=none},
    ytick pos = left,
    xticklabels = {0-2,3-9,10-19,20-29,30-39,40-49,50-59,60-69,70+},
    xticklabel style={rotate=45,anchor=east},
    x label style={at={(axis description cs:0.5,-0.1)},anchor=north},
    title=CLIP Preference for Person Over Gender Labels,
    xlabel= {Age Range},
    legend style={at={(0.65,0.55)},anchor=south west},
    enlarge x limits={abs=1cm}
]

\addplot [pattern=vertical lines,pattern color = blue] coordinates {(1,36.32318181818182) (2,21.28681818181818) (3,8.73409090909091) (4,5.966818181818182) (5,6.030909090909091) (6,6.377727272727272) (7,13.476363636363637) (8,18.609545454545454) (9,36.37727272727273)};

\addplot [pattern=horizontal lines,pattern color = orange] coordinates {(1,61.68363636363636) (2,57.55954545454546) (3,19.762727272727272) (4,11.486818181818181) (5,10.93) (6,10.110454545454544) (7,8.807272727272727) (8,12.805454545454545) (9,31.934545454545454)};

\legend {Black Female, White Male};

\end{axis}
\end{tikzpicture}
\caption{Individuals who are Black and Female are less likely than individuals who are White and Male to receive a label with gender omitted before the age of 20, but are more likely to receive the "person" label (with gender omitted) after the age of 50.}
\label{intersection_gender_by_age}
\Description{Figure depicting CLIP's preference for a person label over an age label, broken down by race, gender, and age.}
\end{figure}
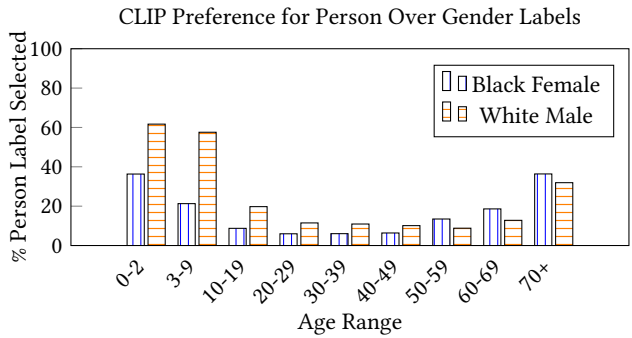

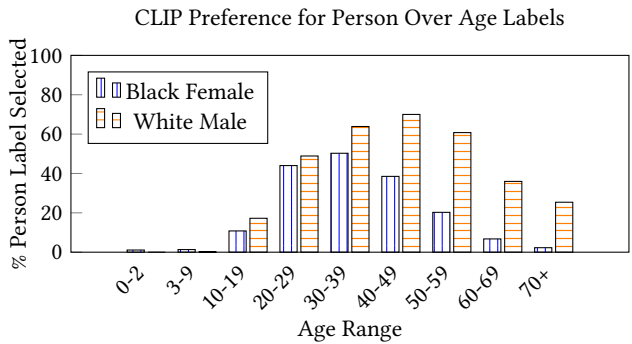
\begin{figure}[!htbp]
\begin{tikzpicture}
\begin{axis} [
    height=4.2cm,
    width=9cm,
    ybar = .05cm,
    bar width = 6.5pt,
    ymin = 0, 
    ymax = 100,
    ylabel=\% Person Label Selected,
    ylabel shift=-5pt,
    ylabel near ticks,
    xtick = {1,2,3,4,5,6,7,8,9},
    xtick style={draw=none},
    ytick pos = left,
    xticklabels = {0-2,3-9,10-19,20-29,30-39,40-49,50-59,60-69,70+},
    xticklabel style={rotate=45,anchor=east},
    x label style={at={(axis description cs:0.5,-0.1)},anchor=north},
    title=CLIP Preference for Person Over Age Labels,
    xlabel= {Age Range},
    legend style={at={(0.03,0.55)},anchor=south west,nodes={scale=1, transform shape}},
    enlarge x limits={abs=1cm}
]

\addplot [pattern=vertical lines,pattern color = blue] coordinates {(1,1.0999999999999999) (2,1.33) (3,10.80090909090909) (4,44.005) (5,50.265454545454546) (6,38.52090909090909) (7,20.248636363636365) (8,6.742272727272727) (9,2.2831818181818178)};

\addplot [pattern=horizontal lines,pattern color = orange] coordinates {(1,0.0) (2,0.16545454545454547) (3,17.23090909090909) (4,48.88136363636363) (5,63.86681818181819) (6,69.99863636363635) (7,60.767272727272726) (8,35.96454545454545) (9,25.382727272727273)};

\legend {Black Female, White Male};

\end{axis}
\end{tikzpicture}
\caption{In every age range after 3-9, individuals who are White and Male are more likely to receive the "person" label (with age omitted) than individuals who are Black and Female. The gender disparity in age labeling observed in Figure \ref{preference_age_by_age} is further exacerbated when the Male group is also White, and the Female group is also Black.}
\label{intersection_age_by_age}
\Description{Figure depicting CLIP's preference for a person label over a gender label, broken down by race, gender, and age.}
\end{figure}

Figure \ref{intersection_gender_by_age} shows that, until the 50-59 age range, individuals who are White and Male are less frequently marked based on gender than are individuals who are Black and Female, with the largest disparities in the 0-2 and 3-9 age ranges. In the 50-59, 60-69, and more than 70 age ranges, individuals who are Black and Female are marked based on gender less frequently than individuals who are White and Male. This is the only circumstance in which individuals who are Black and Female are marked less frequently than individuals who are White and Male, for gender, age, or race or ethnicity labels. Figure \ref{intersection_age_by_age} shows that, in every age range, individuals who are White and Male are less likely to be marked based on age than individuals who are Black and Female. For all age ranges from 30-39 through more than 70, the disparity is similar but more severe than the disparity observed based solely on gender, and is most pronounced at the 50-59 age range (60.8\% vs. 20.2\% unmarked). These results suggest that, as with measurements of self-similarity, already existing biases related to CLIP's preference to mark are magnified in the intersection of race, gender, and age.

\section{Discussion}

Our results provide evidence of biases related to the representation of gender, age, and race or ethnicity in the CLIP embedding space. While prior work has demonstrated semantic biases in CLIP \cite{radford2021learning, agarwal2021evaluating, goh2021multimodal, wang2021gender}, our results indicate a fundamental form of descriptive bias in the model, related to who is marked, and according to what socially defined characteristics they are marked. Representations of Male individuals; White individuals; and individuals between the ages of 20 and 59 are consistently among the least self-similar, and the least likely to be marked. Representations of individuals who are Female; who belong to underrepresented racial or ethnic groups; and who are under the age of 20 or over the age of 60 are consistently the most self-similar, and the most likely to be marked.

Differentially greater self-similarity is not a desirable property in a multimodal embedding space, as it reflects that the embeddings cluster more tightly around the gender, age, or race or ethnicity of the people depicted in the images. The disparities between the most self-similar and least self-similar social groups suggests that independently impactful biases based on race, gender, and age compound to multiply impact groups possessing more than one of these constituent characteristics. Results based on self-similarity are most comparable to prior work finding that low-frequency names which are statistically more associated with women and with underrepresented races and ethnicities are more self-similar across contexts in language models, and are represented using a smaller region of the embedding space than names more associated with men and with people who are White \cite{wolfe2021low}. Our results indicate that such effects are not limited to linguistic representations, but extend to language-and-vision embedding spaces wherein representations have conceptual features. Moreover, using a large dataset of faces overcomes the ambiguity of measuring the bias of a linguistic representation like a name. 

Our results demonstrate the effects of natural language supervision in shaping a multimodal embedding space. Among the most consistent results is that age has significant impact on whether the model prefers to mark not only age but also gender and, to a lesser extent, race or ethnicity. Among Female individuals, 91.5\% of those between 30 and 39 are marked based on gender, compared to 60.8\% of those more than 70. This suggests that the model arranges its embedding space based not on fine-grained recognition of visual information, but on the categories used to describe similar images in the training data. Moreover, such results suggest that age should be considered as a factor in interpreting the results of research on bias in AI, even when that research does not intend to test primarily the effects of age.

The results of the markedness experiment appear to form patterns similar to a normal distribution across age ranges, with a mean centered at 30-39, and shifted based on gender or on race or ethnicity. Given that this experiment quantifies how frequently "a photo of a person" was selected over more descriptive labels, the normality of these distributions suggests that CLIP represents certain characteristics (White, Male, between 20 and 59) as more central to the concept of "person," and other characteristics as different enough that they need to be marked using more descriptive language.

\noindent\textbf{Limitations and Future Work.} Our research uses the FairFace dataset to study bias in CLIP. While this is the most balanced and diverse dataset of human images of which we are aware, it is limited in that it uses a small number of socially defined categories for gender and for race or ethnicity, and assigns categories based not on self-identification but on the perception of annotators. In adopting this dataset as a source of images, our results are necessarily contextualized within those categories. Diverse, balanced, and ethically obtained datasets of human images are needed to permit less constrained research designs at a similar scale (more than $80,000$ images) to our work. Our research examines only one visual semantic model, and future work will be needed to assess whether our results generalize across architectures. As the field of language-and-vision AI matures, systematic study across architectures may reveal additional insight into the bias of markedness. Moreover, CLIP embeds highly contextual sentence representations, and the results of the markedness experiment may be affected by using different prompts or social categories. We have adopted a principled approach using only the prompt specified by \citet{radford2021learning} and the social categories defined by \citet{karkkainen2021fairface}. Future work might explore how systematically varying such settings affects CLIP's preference to mark.

\citet{radford2021learning} note that CLIP was first intended to have the capabilities of a zero-shot caption generator, \textit{i.e.,} a model capable of not just matching an image with a label but of producing that caption using a language model \cite{radford2021learning}. Moreover, one of the first uses of CLIP was to train DALL-E, a zero-shot text-to-image transformer capable of generating novel visual representations from text input \cite{ramesh2021zero}. As such generative models are developed, research might be directed to understanding what socially defined categories are made conspicuous or invisible in the text and images generated by AI, in ways which could impact society. Finally, future work might explore the connection between self-similarity and frequency of representation in multimodal training corpora. Prior work suggests that low frequency leads to high self-similarity in language models \cite{wolfe2021low}, and our results identify a similar effect in multimodal AI.

\section{Conclusion}
We demonstrate that a multimodal visual semantic AI model learns to unequally mark gender, age, and race or ethnicity. Biases learned in visual semantic embedding spaces are likely to affect the next generation of state-of-the-art AI applications, which build on the ability of such models to associate images with text in a zero-shot setting.

\begin{acks}
This material is based on research partially supported by the U.S. National Institute of Standards and Technology (NIST) Grant 60NANB\-20D212T. Any opinions, findings, and conclusions or recommendations expressed in this material are those of the authors and do not necessarily reflect those of NIST.
\end{acks}

\bibliographystyle{ACM-Reference-Format}
\bibliography{sample-base}

\end{document}